\documentclass[12pt,noinfoline]{imsart}

\usepackage[margin=0.75in]{geometry}

\RequirePackage[OT1]{fontenc}
\usepackage{amsthm,amsmath,natbib}
\RequirePackage[colorlinks,citecolor=blue,urlcolor=blue]{hyperref}

\usepackage{blindtext} 
\usepackage{float}

\usepackage[normalem]{ulem}
\usepackage{bm}
\usepackage[english]{babel} 
\usepackage{graphicx}
\usepackage{amsmath}
\usepackage{amssymb}
\usepackage{mathtools}
\usepackage{tikz}
\usetikzlibrary{arrows,decorations.pathmorphing,backgrounds,positioning,fit,petri}
\usetikzlibrary{shapes}
\usepackage{dirtytalk}
\usepackage{hyperref} 

\startlocaldefs
\numberwithin{equation}{section}
\theoremstyle{plain}

\DeclareMathOperator*{\argmin}{argmin} 
 
\DeclareMathOperator*{\median}{median} 

\endlocaldefs

\begin{document}

\begin{frontmatter}
\title{A Bayesian  Perspective  of Statistical  Machine  Learning  for
Big Data} 
\runtitle{Bayesian Machine Learning}

\begin{aug}
\author{\fnms{Rajiv} \snm{Sambasivan,} \ead[label=e1]{rajiv.sambasivan@gmail.com}}
\author{\fnms{Sourish} \snm{Das}\ead[label=e2]{sourish@cmi.ac.in}
\ead[label=usd,url]{https://www.cmi.ac.in/~sourish/}}
\and
\author{\fnms{Sujit K.} \snm{Sahu}
\ead[label=e3]{S.K.Sahu@soton.ac.uk}
\ead[label=u1, url]{http://www.soton.ac.uk/$\sim$sks/}
}
\runauthor{Sambasivan et al.}

\affiliation{Chennai Mathematical Institute, India and University of Southampton, UK}

\address{\printead{e1},}
\address{\printead{e2},}
\address{\printead{e3}.}
\end{aug}

\begin{abstract}
  Statistical Machine Learning (SML) refers to a body of algorithms and methods by which computers are allowed to discover important features of input data sets which are often  very large in size. The very  task of feature discovery from data is essentially the meaning of the keyword `learning' in SML. Theoretical  justifications for the effectiveness of the SML algorithms are underpinned by sound principles from different   disciplines, such as Computer Science and Statistics. The  theoretical underpinnings particularly justified by statistical inference methods are together termed as  statistical learning theory.

  This paper provides a review of  SML from a Bayesian decision theoretic point of view --  where we argue that many SML  techniques are closely connected to making inference by using the so called  Bayesian paradigm. We discuss many important SML techniques such as supervised and unsupervised learning, deep learning, online learning and Gaussian processes especially in the context of very large data sets where these are often employed. We present a  dictionary which maps the key  concepts of SML from Computer Science and Statistics. We illustrate the SML techniques with three moderately large data sets where we also discuss many practical implementation issues. Thus the review is especially targeted at statisticians and computer scientists who are aspiring to understand and apply SML for moderately large to big data sets. 

\end{abstract}


\end{frontmatter}

\section{Introduction}\label{sec:intro}

Recently, there has been an  exponential increase in the number of smart devices, known as ``Internet of Things" (IoT), that connect to the internet, and consequently, massive data sets are generated by the users of such devices.  This has led to a  phenomenal increase in operational data for many  corporations, businesses, and government organizations. Clearly, each of these organisations now sees a plethora of opportunities in learning important features, regarding consumer behaviour among other aspects, from such large data sets, which can be translated into actionable insights collectively known as business intelligence. Business intelligence is an umbrella term to describe the processes and technologies that are used by organizations to leverage their large data sets to create profit making acumen. Many companies and organisations, such as Google and Facebook, have enormously benefitted by using SML techniques, which in turn justify the tremendously growing interest in the statistical theory behind SML. 

The explosion of operational data, collected from IoTs or otherwise, poses technological challenges to store data in data repositories and warehouses which are  sometimes called \emph{data lakes}, see e.g. \cite{kimball1998data} and \cite{inmon2016data}.  Often, organizations leverage the data stored in these data lakes using two most common approaches. The first is a top-down approach where a user has a specific query or hypothesis to test. For example, a retail company might be interested in finding an additional expense of \$1000 on digital marketing is going to boost the revenue significantly. Such queries or hypotheses are formulated a priori before starting the investigation. This type of analysis works when users exactly know what to look for in the data. The second approach is used to discover insights that the users have not explicitly looked to determine but would like to discover from the data alone.  This is called \emph{data mining} and the most common approach to  data mining is \emph{machine learning} which is characterized by its use of statistical theory and methods. Recently, many researchers are using an alternative mathematical data exploration approach called topological data analysis, see e.g. \cite{holzinger2014topological}. This paper, however, does not discuss those approaches.

Motivated by the enormous success of SML techniques for large data sets we set out to find accessible reviews of these techniques in the literature. Indeed, there are many such reviews and a few notable ones include:  \cite{domingos2012few,rao.govindaraju.2013, al2015efficient, friedman2001elements} and  \cite{Qiu.survey.ML.2016} and the references therein. \cite{domingos2012few} discusses the general area of machine learning but does not focus on SML. The book by \cite{rao.govindaraju.2013} covers many theoretical aspects, such as sequential bootstrap, cross-entropy method, bagging, boosting and random forest method.  The article by \cite{al2015efficient}  discusses the challenges of big data and some techniques to scale machine learning to big data sets from a computer science perspective. The book by \cite{friedman2001elements} focuses on a decision theoretic approach to SML. However, the book does not discuss many recently successful  concepts, such as the Gaussian process prior models.  Finally, \cite{Qiu.survey.ML.2016} present an exhaustive literature survey, but they  do not provide a review of statistical learning theory.

The main contribution of this  paper is to present many  SML techniques by using a Bayesian decision theoretic framework following along the lines of \cite{friedman2001elements} but including more recent concepts of Gaussian process prior models and techniques such as deep learning for big data. Aiming to be  a bridge between the machine learning literature and statistical decision theory, this paper provides a dictionary which maps the key  concepts of SML from Computer Science and Statistics. 
The paper  also  discusses model complexity and penalization methods using arguments based on Bayesian methods. The SML techniques are illustrated with three moderately large data sets where we also discuss many practical implementation issues. We also discuss machine learning techniques suitable for big data.

The rest of this paper is organized as follows. In Section~\ref{sec:sml}, we present the main ideas of SML and  discuss the nature of typical problems solved using  SML.
In Section~\ref{sec:sml} we also present a dictionary, which maps similar concepts used in Statistics and Computer Science. In Section~\ref{sec:bd}, we present a description of big data and its characteristics with a view to applying  SML to such data sets. Section~\ref{sec:slt} provides an overview of theoretical ideas needed for the analysis of SML algorithms. In Section~\ref{sec:mlbd}, we discuss the challenges of applying machine learning to big datasets and the computational approaches to address these challenges. Section~\ref{sec:mlbd} describes the application of SML to big data using the ideas based on statistical learning theory previously presented in  Section~\ref{sec:slt}.  In Section~\ref{sec:spiml} we discuss some important practical issues in applying SML.  Section~\ref{sec:application} presents a comparative study of different methods presented in Sections~\ref{sec:slt}~and~\ref{sec:mlbd} with three different datasets.  Finally, we conclude this work with a few summary remarks in Section~\ref{sec:summary}.


\section{The Main Keywords in Statistical Machine Learning}\label{sec:sml}

The major concepts of SML have been in the purview of both Computer Scientists and Statisticians for quite a while. SML is an outcome of the natural intersection of both Computer Science and Statistics, see \cite{mitchell2006discipline} and \cite{bsm_website}. Statisticians and Computer Scientists often use different terminologies to describe the same idea and method. For example, the attributes in a dataset are called \textbf{\emph{covariates}} by the Statistical community and \textbf{\emph{features}} by the Computer Science community.  In  Table \ref{tbl_dictionary} we present a dictionary which maps the key concepts between the Computer Science and Statistics, see \cite{wasserman2013all}[Preface] for more details about the table. In the remainder of this  section, we describe  the main keywords in SML.\\

\begin{table}[ht!]
\caption{Dictionary of Same Concepts between Statistics and Computer Science}\label{tbl_dictionary}
\begin{tabular}{|c|c|l|}\hline
\textbf{Statistics} & \textbf{Computer Science} & \textbf{Explanations} \\ \hline
Data    & Training or In-sample & $(x_1,y_1),\ldots, (x_n,y_n)$ \\
&& the data to fit/train the model\\ \hline
- & Test or Out-sample & $(x_{n+1},y_{n+1}),\ldots, (x_{n+m},y_{n+m})$ \\
&& the data to test the accuracy of prediction \\
&& from the trained or fitted model\\ \hline
Estimation & Learning & use data to estimate (or learn)  unknown quantity \\
&& or parameters of the model \\ \hline 
Classification & Supervised Learning & predicting a discrete $y$ from $X$ \\ 
&&\\\hline
Regression & Supervised Learning & predicting a continuous $y$ from $X$ \\ 
&&\\\hline
Clustering & Unsupervised Learning & putting data into groups\\
&&\\ \hline
Dependent variable & Label (or Target) & the $y_i$'s \\ &&\\ \hline
Covariates or Predictors & Features & the $X_i$'s \\ &&\\ \hline
Classifier & Hypothesis & map from covariates to outcome \\ 
&& \\ \hline
Hypothesis & - & Subset of parameter space $\Theta$, which is supposed to be true \\
&& \\ \hline
Confidence interval & - & Interval that contains an
unknown quantity \\
&& with certain probability \\
&& \\ \hline
Bayesian inference & Bayesian inference & statistical methods for using data to update probability\\
&& \\ \hline
frequentist inference & - & statistical methods
with guaranteed frequency behavior\\ 
&& \\ \hline
Directed acyclic graph & Bayesian net & multivariate distribution with conditional \\
&& independence relations \\
&& \\ \hline
Statistical Consistency  & PAC learning & uniform bounds on
probability of errors \\ 
Large deviation bounds&&\\
&& \\ \hline
- & semi-supervisd learning & limited amount of labeled data is used \\
&& in combination with unlabeled data to \\
&& perform the learning task. \\
&& \\\hline
stochastic games & reinforcement learning & an agent interacting with its environment\\
&& to maximize reward \\
&& \\ \hline
Sequential Analysis & On-line learning & Receives data sequentially. \\
&& learn or predict an incoming stream of observations,\\
&& one sample
at a time.\\ \hline
Class of Models ($\mathcal{M}$) & Hypothesis Class ($\mathcal{H}$) & Set of models like logistic regression\\
&& for binary classification problem\\ \hline
\end{tabular}
\end{table}

\noindent \textbf{Learning}: What is known as \emph{parameter estimation} or estimation of \emph{unknown functions} in Statistics, is known as ``\textbf{learning}" in Computer Science. That is, SML is concerned with learning from data. 
\textbf{Supervised Learning}:
A  categorization of learning tasks  based on the use of a label, (also known as the target or dependent variable) to perform the learning activity. Tasks that use labels to perform the learning activity are called \emph{supervised learning} tasks. The label can be either a discrete or a continuous quantity. Supervised learning tasks where the label is a discrete quantity are called \emph{classification} tasks. When the label is continuous, the learning task is called \emph{regression}. The goal of a learning task associated with a label is to predict it.  For example, an application that performs fraud detection based on transaction characteristics is an example of supervised learning. In another example, credit card companies heavily use supervised learning to identify good customers, where they use the potential customers demographic profile and credit history as covariates or features in their classification task.

\textbf{Unsupervised Learning}: Not all learning tasks are associated with labels. Such tasks are called \emph{unsupervised learning} tasks. These tasks do not use a label to perform the learning activity. An essential problem in unsupervised learning involves grouping  similar data points. This task is called \emph{clustering}. During exploratory analysis of the data for a learning task, analysts often use techniques that express the observed variability regarding a small number of uncorrelated unobserved factors. Reducing the number of variables helps in understanding the characteristics of the data and the problem. This technique is called \emph{factor analysis} and is another example of an unsupervised learning task. Unsupervised learning has many practical applications, see e.g., \cite{friedman2001elements}[Chapter 14] for more details.

\textbf{Semi-supervised learning}: There are application areas where labeled data are scarce, and a limited amount of labeled data is used in combination with unlabeled data to perform the learning task. This kind of learning is called \emph{semi-supervised learning}, see e.g., \cite{ssl_2006}. Semi-supervised learning has claimed great successes in image classification and text processing applications, see e.g., \cite{Guillaumin.2010}. 

\textbf{Active learning}: Another type of learning that bears a strong connection with semi-supervised learning is \emph{active learning}. As with semi-supervised learning, the labeled data is a scarce commodity in active learning. However, in active learning, the learning algorithm can interact with an oracle (who may be a human annotator) to pick this set of labeled data. A successful application of active learning is in the area of information retrieval, see e.g., \cite{settles2012active}.

\textbf{Reinforcement learning}: Another category of learning, called \emph{reinforcement learning}, characterizes the learning task as an agent interacting with its environment to maximize reward, see e.g., \cite{sutton1998introduction}. The agent learns how to map situations (an operational context) into actions so that it can have the maximum reward. Such a learning framework is a natural fit for many application areas, for example, online advertisement, robotic control, recommender systems, and video games. The Markov decision process from Statistics is an important tool that is used in reinforcement learning, see e.g., \cite{Littman.1994}.

Bayesian reinforcement learning explicitly elicits prior distribution over the parameters of the model, the value function, the policy, and its gradient. It then updates the Bayes estimators, each time it receives feedback, see e.g., \cite{MAL-049} and \cite{Vlassis.2012}. An analytic solution to discrete Bayesian reinforcement learning is presented in \cite{Poupart.2006}.

\textbf{Transfer learning}: Most machine learning applications fall into the categories discussed above. In addition to these, there exist some other niche types of learning. When confronted with a problem for which the data is limited, we may be able to leverage the learning results from a similar problem. This has been  successful in some application areas like image classification and small area estimation. This type of learning is called \emph{transfer tearning}, see e.g., \cite{pratt1993discriminability} and \cite{torrey2009transfer} for an overview.

\textbf{Inductive Learning and Transductive learning} In many applications of machine learning, we hope to generalize the results. In other words, we would like the results obtained from performing the learning task to apply to a new data set. This kind of learning is called \emph{inductive learning}. We discuss this in Section \ref{sec:slt}. In some applications, we may not need this generalization, where we are only interested in performance on specific data, see e.g.,  \cite{gammerman1998learning}. Learning that can achieve this goal is called \emph{transductive learning}, see e.g., \cite{gammerman1998learning} and \cite{pechyony2009theory} for details. Transductive learning has been applied to solve problems on graphs and in text mining, see e.g., \cite{Joachims.1999}.

\section{Statistical Machine Learning for Big Data}\label{sec:bd}
SML applications today need to perform over big data. Big data is a term used by practitioners to describe data with characteristics that make it difficult for conventional software packages and infrastructure to process. There is an effort to standardize the definition of big data, see \cite{nist_bd_defn}[Section 2.1]. The characteristics of big data are:
\begin{enumerate}
\item \textbf{Volume}: To learn insight, the processing of the high volume of operational data is important.
\item \textbf{Variety}: Operational data may arrive from multiple systems each with its own set of characteristics.
\item \textbf{Velocity}: Operational data may arrive at high speed.
\item \textbf{Variability}: The meaning of the same data could vary over time. The variability is common in the data from applications that have natural language processing capabilities. The same words tend to mean different things in different contexts.
\end{enumerate}

SML applications operating on big data have had to adapt to these challenges. Implementation of SML is often tightly coupled to new computational frameworks for computation on big data. Widely used computational frameworks include:
\begin{itemize}
\item \textbf{MapReduce}: It is a programming paradigm, associated with implementation for processing big data sets with a parallel, distributed algorithm on clusters of computers, see e.g.,  \cite{dean2008mapreduce} and \cite{leskovec2014mining}[Chapter 2] for further details.
\item \textbf{Graph Lab}: This is another high performance, distributed computation framework written in C++.  It is a graph-based approach to computation on big data, originally developed in Carnegie Mellon University, see e.g., \cite{graph_lab}.
\item \textbf{Spark}: This is a distributed general-purpose cluster-computing, open source framework, similar to Graph Lab. It was developed at the University of California, Berkeley, see \cite{zaharia2016apache}.
\end{itemize} 
It should be evident that apart from statistical theory, ideas from Computer Science, such as algorithms, data structure and distributed systems are needed to implement SML applications. The term data used in the description of big data in the above discussion refers to raw data that may not be in the form required for SML. Data used by SML applications typically have a tabular representation. Significant processing effort is needed to convert raw operational data into the tabular representation used by SML algorithms. It is not hard to see that digital technology, like IoT, social media etc. has resulted in big datasets. From the standpoint of SML algorithms processing big data, there is another perspective to the term ``\textbf{big}." The dimensions of the tabular representation affect the computation performed by SML algorithms. The height of the tabular representation provides the number of tuples, $n$,  or sample size, and the number of values in the tuple, $p$, or variables. 

%

%
The problem of performing Statistical machine learning (SML) over data with a large sample size of $n$  refers to as the \say{big n} problem. The problem of performing SML over data with a large number of covariates $p$ is called the \say{big p} problem. Problems that are big regarding features and are big concerning the sample sizes are now emerging in some application areas. See \cite{ravi_kumar_statistical_nodate} for a discussion of these problems. Techniques to work on both \say{big n} and \say{big p}  and many other several challenges to machine learning on big data sets that are presented in Section \ref{sec:mlbd}.  Before that, we introduce the theoretical framework of statistical learning in the following Section.

\section{Theoretical Framework for Statistical Learning} \label{sec:slt}

Statistical Learning theory provides the concepts and the analytical framework for the analysis of machine learning algorithms. For a historical perspective of the development of Statistical Learning,  see \cite{vapnik1998statistical}[Chapter 0]. In this section, we summarize the essential conceptual ideas used in statistical learning. The goal of statistical learning theory is to study, in a statistical framework, the properties of learning algorithms, see \cite{bousquet2004introduction}. The statistical learning theory is rooted in the Bayesian decision theoretic framework, see \cite{Berger.1993}.

\subsection{Bayesian Decision Theoretic Framework}

The central problem of statistical learning theory is the estimation of a function from a given set of data. The notation used to formalize the problem is as follows. The data for the learning task consist of attributes $x$ with labels $y$. The input space is $\mathcal{X}$, and the output space is $\mathcal{Y}$. When the learning task is classification, the output space is finite. For example, in a binary classification problem,  $\mathcal{Y} = \{1, -1\}$. When the learning task is the regression, the output space is infinite, $\mathcal{Y} \subseteq \mathbb{R}$. The data set $D$ consists of $(x, y)$ pairs  from an \emph{unknown} joint distribution $P(x,y)$. We seek a function $f(x)$ for predicting $y$ given values of the input $x$. The goal of learning is to learn the function $f: \mathcal{X} \mapsto \mathcal{Y}$ that can predict the label $y$ given $x$. We cannot consider all possible functions. We need some specification of the class of functions we want to consider for the learning task.

The class of functions considered for the learning task is called the \emph{hypothesis class or class of models}, $\mathcal{F}$, see Table \ref{tbl_dictionary}. Consider a function $f \in \mathcal{F}$. The hypothesis class $\mathcal{F}$ could be finite or infinite. An algorithm $\mathcal{A}$ is used to pick the best candidate from $\mathcal{F}$ to perform label prediction. To do so,  $\mathcal{A}$ measures the \emph{loss} function $L(f(x), y)$, a performance indicator for $f(x)$. It measures the loss due to the error in the prediction from $f(x)$ against the true label $y$.   The loss, $L(f(x), y)$, is a random variable. Therefore we need the expected value of the \emph{loss} to characterize the performance of $f$. This expected value of the loss is called the \emph{risk} and is defined as:
\begin{equation}\label{eqn:risk}
R(f) = \mathbb{E}[L]=\int L(f(x),y)P(dx,dy).
\end{equation}
By conditioning on $x$, we can write Equation (\ref{eqn:risk}) as
\begin{equation*}
R(f) = \mathbb{E}[L]=\mathbb{E}_{x}\mathbb{E}_{y|x}(L(f(x),y)|x),
\end{equation*}
and the $R(f)$ is also known as the posterior expected loss, see \cite{Berger.1993}[Chapter 4]. It is sufficient to minimize the risk $R(f)$ point wise:
\begin{equation}\label{eqn_post_expect_loss_cont}
f(x)=\argmin E_{y|x}(L(f(x),y)|x).
\end{equation}
The solution $f(x)$ in (\ref{eqn_post_expect_loss_cont}) is known as the Bayes estimator under loss function $L$, see \cite{Berger.1993}[Chapter 1 and Chapter 4] for more detail. When $y$ assumes the values in $\mathcal{Y}=\{1,2,\cdots,K\}$, i.e., the set of $K$ possible classes. The loss function $\mathbf{L}$ can be presented as a $K \times K$ matrix. The $(k,j)^{th}$ elements of the loss matrix $\mathbf{L}$ is
\begin{eqnarray*}
L(k,j)=\Big \{ \begin{array}{cc}
0 & k=j \\
l_{kj} & k \neq j,
\end{array}
\end{eqnarray*}
where $l_{kj}\geq 0$ is the penalty for classifying an observation $y_j$ wrongly to $y_k$. A popular choice of $\mathbf{L}$ is the zero-one loss function, where all misclassification are penalized as single unit. We can write the risk as
\begin{equation*}
R(f)=\mathbb{E}_{x}\sum_{k=1}^{K}L[y_k,\hat{f}]P(y_k|x),
\end{equation*}
and it suffices to minimize $R(f)$ point wise:
\begin{equation*}
R(f)=\argmin_{f \in \mathcal{Y}}\sum_{k=1}^{K}L[y_k,\hat{f}]P(y_k|x).
\end{equation*}
With the $0-1$ loss function this simplifies to
\begin{equation*}
f(x)=	\max_{y \in \mathcal{Y}} P(y|x).
\end{equation*}
This solution is known as \emph{Bayes classifier}, see \cite{Berger.1993} and  \cite{friedman2001elements}[Chapter 2]. This method classifies a point to the most probable class, using the posterior probability  of $P(y|x)$. The error rate of the Bayes classifier is known as the \emph{Bayes rate} and the decision boundary corresponding to Bayes classifier is known as the Bayes-optimal decision boundary. Suppose $p_k(x)$ is the class-conditional density of $x$ in class $y=k$, and let $\pi_k=P(y=k)$ be the prior probability of class $k$, with $\sum_{k=1}^{K}\pi_k=1.$ A simple application of Bayes theorem gives us
\begin{equation*}
P(y=k|x)=\frac{p_k(x)\pi_k}{\sum_{l=1}^{K}p_l(x)\pi_l}.
\end{equation*}
We see that in terms of ability to classify, having the $p_k(x)$ is almost equivalent
to having the posterior probability of $P(y=k|x)$. Many techniques are based on models for the class densities: (1) linear and quadratic discriminant analysis use Gaussian densities; (2) general nonparametric density estimates for each class density allow the most flexibility; (3) Naive Bayes models are a variant of the previous case, and assume that each of the class densities are products of marginal densities; i.e., they assume that the inputs are conditionally independent in each class.

When $y$ is continuous, as in a regression problem, the approach discussed in Equation (\ref{eqn:risk}) and (\ref{eqn_post_expect_loss_cont}) works, except that we need a suitable loss function for  penalizing the error. The most popular loss function is squared error loss: $L(f(x),y)=(y-f(x))^2$ and the solution under the loss function is
$$
f(x)=E(y|x),
$$
the conditional expectation also known as the \emph{regression} function. If we replace the squared error loss by the absolute error loss, i.e., $L(f(x),y)=|f(x)-y|$, then the solution is the conditional median, 
$$
f(x)=\median(y|x),
$$
its estimates are more robust than those for the conditional mean.

\subsection{Learning with Empirical Risk Minimization}\label{subsec_ERM}

The learning of $f$ is performed over a finite set of data often called the \emph{training dataset}. To evaluate the expected loss in Equation (\ref{eqn:risk}), we need to evaluate the expectation over all possible datasets. In practice, the joint distribution of the data is unknown, hence evaluating this expectation is intractable. Instead, a portion of the dataset $D_{training} \subset D $, is used to perform the learning and the remainder of the dataset $D_{test} = D - D_{training}$, is used to evaluate the performance of $f$ using the loss function $L$. The subset of the data used for this evaluation, $D_{test}$, is called the \emph{test dataset}. The expected loss over the training dataset is the \emph{empirical risk} and is defined as:
\begin{equation}
\label{eqn:emp_risk}
R(\hat{f}) = \frac{1}{n}\sum_{i = 1}^{ n} L(f(x_i), y_i).
\end{equation}
Here $n$ represents the number of samples in the training dataset, $D_{training}$. The learning algorithm uses the empirical risk, $R(\hat{f})$, as a surrogate for the true risk, $R(f)$, to evaluate the performance of $f$.  The best function in $\mathcal{F}$ for the prediction task is the one associated the lowest empirical risk. This principle is called \emph{Empirical Risk Minimization}, and is defined as\\
\begin{equation}\label{eqn:min_emp_risk}
h = \inf_{f \in \mathcal{F}} R(\hat{f}).
\end{equation}

The task of applying algorithm $\mathcal{A}$ to determine $h$ is called the \emph{learning task}. Implementation of algorithm $\mathcal{A}$ is called the \emph{learner}. For example, the maximum likelihood procedure is an example of the \emph{learner}. The lowest possible risk for the learning problem, $R^*$ associated with the function $f^*$,  is obviously of interest to us. The lowest possible risk is the \emph{Bayes risk}. The hypothesis class for $f^*$ may or may not be the same as $\mathcal{F}$. Consider the output produced by the learner $h$. Comparing the risk for $h$ with $R^*$ (\emph{excess risk}) provides another perspective of the quality of $h$, presents in Equation  (\ref{eqn:excess_risk_decomp}):
\begin{equation}\label{eqn:excess_risk_decomp}
R(\hat{h}) - R^* = \left(R(\hat{h}) - R(h) \right) + \left( R(h) - R^*\right).
\end{equation}
In Equation (\ref{eqn:excess_risk_decomp}), the first term is the \emph{estimation error} and the second term is the \emph{approximation error}. Consider the hypothesis class that represents hyperplanes. The algorithm estimates the parameters of the hyperplane from the training dataset using the \emph{empirical risk minimization}. If we use a different training dataset, then the estimates for the hyperplane are different, and such error arises from using a sample to learn $h$. Accordingly, it is called the estimation error. The approximation error indicates how well a hypothesis class can approximate the best function. Let us consider an example where $f^*$ (associated with $R^*$) is complex, say a high degree polynomial and our hypothesis class $\mathcal{F}$ is the set of hyperplanes. We should expect to see a high approximation error. Note that we observe the high approximation error, if $f^*$ is simple (like a hyperplane) and our hypothesis class $\mathcal{F}$ is complex (like a class of high degree polynomials). It is evident that the choice of the hypothesis class $\mathcal{F}$ affects the performance of the learning task. In particular, the complexity associated with the hypothesis class $\mathcal{F}$ can influence the level of risk that the learning algorithm $\mathcal{A}$ can achieve. 

In summary, both the approximation error and the estimation error affect the performance of the algorithm to determine the best hypothesis class for a problem. One must make a trade-off between the two, and it is called the \emph{bias-variance} trade-off, see  \cite{friedman2001elements}. Choosing hypothesis classes that are more complex than what is optimal can lead to a phenomenon called \textbf{\emph{over-fitting}}. Often, over-fitting implies very good performance of the class on the training data set but very poor performance on the test data. The capability of the function determined by algorithm $\mathcal{A}$ to maintain the same level of precision on both the training and test datasets is called \textbf{generalization}. If the algorithm $\mathcal{A}$ generalizes well, then the new insight learned from modeled data is likely to be \textbf{reproducible} in the new dataset, provided the training dataset is true representation of the population.

\subsection{Bayesian Interpretation of the Complexity Penalization Method}\label{subsec_penalization_method}

Datasets with complex structure occur in many applications. Using a complex hypothesis class on a simple learning problem and simple hypothesis class on a complex problem, both result in poor performance.  Hence we need methods that are sophisticated to handle the complexity of the problem. The method should consider a set of hypothesis, and pick an optimal hypothesis class based on an assessment of the training data.  The method must also achieve good generalization and must avoid over-fitting. A class of methods that can achieve good generalization are known as \emph{Complexity Penalization Methods}. These methods include a penalty for the complexity of the function while evaluating the risk associated with it,  see \cite{friedman2001elements}  for details. The general template to determine the solution $h$ of complexity penalization method is:
\begin{equation}\label{eqn:comp_reg}
h = \argmin_{f \in \mathcal{F}} \big\{\hat{R}(f) + C(f)\big\}, 
\end{equation}
where $C(f)$ is the term associated with the complexity of the hypothesis class $\mathcal{F}$. The solution $h$ in (\ref{eqn:comp_reg}) is the solution of constrained optimization of the risk $\hat{R}(f)$, where $C(f)$ is the cost or constrained on $\hat{R}(f)$.

We have not yet discussed ways of specifying the complexity $C(f)$ of the hypothesis class. There are many available methods and the right choice depends on the learning problem and hypothesis class. 
The intent here is to point out that methods to specify the complexity of the hypothesis class exists. Examples of choices used to specify the complexity of the hypothesis class include \emph{VC dimension (Vapnik Chevronenkis dimension)}, \emph{Covering Number} and \emph{Radamacher Complexity}, see \cite{bousquet2004introduction} for details.

The Bayesian approach has one-to-one correspondence with the complexity penalization method. In a Bayesian approach, we consider a probability distribution over $\mathcal{F}$. The Bayes rule can determine the probability of a particular model as,
\begin{equation}\label{eqn:bayes_rule}
P(f|x) = \frac{P(x|f) P(f)}{P(x)},
\end{equation}
where $x$ represents the input. The denominator in Equation (\ref{eqn:bayes_rule}) is the normalizing constant to ensure that probabilities associated with the functions in $\mathcal{F}$ integrate to $1$ and the denominator is free from $f$. So after taking the $\log$ on both sides, the Equation (\ref{eqn:bayes_rule}) can be expressed as 
$$
\log(\mathbb{P}(f|x))\propto \log(\mathbb{P}(x|f))+\log(\mathbb{P}(f)).
$$
Consider the right hand side of Equation (\ref{eqn:comp_reg}). The first term, $\hat{R}(f)$ called the risk, is proportional to the negative log-likelihood of the function $f$, i.e., $-\log(\mathbb{P}(x|f))$. The second term of Equation (\ref{eqn:comp_reg}), $C(f)$ can be interpreted as the negative log-prior distribution, i.e., $-\log(\mathbb{P}(f))$ for the problem under consideration, see \cite{friedman2001elements}. The $C(f)$ can also be viewed as a cost function. The cost $C(f)$, is large when the function $f$ is less likely and is small when $f$ is more likely. The solution $h$ in Equation (\ref{eqn:comp_reg}) is the posterior mode of the posterior distribution of $f$, i.e., 
\begin{equation}\label{eqn_post_mode}
h= \argmin_{f \in \mathcal{F}} \big\{-\log(\mathbb{P}(f|x))\big\}.    
\end{equation}
\textbf{Posterior mode} is the Bayes estimator under Kullback-Libeler type loss function, see \cite{Das.Dey.2010}. The posterior mode is estimated via optimization routine. 
In Section \ref{subsub_post_mode_large_opt}, we discuss  the large scale optimization on big dataset to find posterior mode.
The \textbf{posterior mean} and \textbf{posterior median} can also be estimated via the Markov Chain Monte Carlo (MCMC) simulation techniques such as the Gibbs sampler, see \cite{gelfand}. MCMC-based Inference of big data is an active area of research, see e.g., \cite{Rajaratnam.2015}.

\subsection{No Free Lunch Theorem of Statistical Learning}

An important result related to picking a hypothesis class for a problem is the \emph{\say{No Free Lunch Theorem for Statistical Learning}}, see \cite{wolpert1996lack}. The theorem says that there does not exist a single hypothesis class that works for \emph{all} learning problems. Practitioners should keep this result in mind when applying learning theory to practical problems. 
In practice, selecting a good hypothesis class comes from prior experience with a learning problem and exploratory analysis to validate expert judgment. Empirical risk minimization obtains the parameters of the hypothesis class. Having sufficient data ensures that we get good parameter estimates. In summary, good performance on a learning task is achieved by having prior experience for the problem and sufficient data to estimate parameters. Some of these ideas have been used in unsupervised learning as well, see \cite{friedman2001elements}[Chapter 14] for details.

\subsection{PAC Learning and Statistical Consistency}

The idea of \emph{Probably Approximately Correct}  (PAC) learning is parallel to the \emph{asymptotic consistency} of the estimator in Statistics, see e.g., \cite{Haussler.1994}. Recall that the learning task performs with a particular dataset $D$, discussed in Section \ref{subsec_ERM}. The hypothesis class $\mathcal{F}$ considered by the learning algorithm may or may not correspond to the hypothesis class associated with the Bayes risk, $R^*$. Therefore the posterior mode $h$ in Equation (\ref{eqn_post_mode}) determines as the best predictor by the learning algorithm could be an approximation to $f^*$ associated with the Bayes risk. As a consequence, we expect to see some difference ($\epsilon$) between the empirical risk $R(\hat{h})$ and the Bayes risk $R^*$. The empirical risk is determined by a training dataset. A different training data set would yield a different value for the empirical risk associated with the posterior mode $h$. Therefore the quantity $ R(\hat{h}) - R^*$ is a random variable. We can evaluate probability to characterize this variation. This characterization takes the form for a given $\epsilon>0,$ \begin{equation}\label{eqn:pac}
P\left(|R(\hat{h}) - R^* |\leq \epsilon \right) \geq 1 -\delta, 
\end{equation}
where $0 \leq \delta \leq 1$. For a  hypothesis class $\mathcal{H}$, we can evaluate its suitability to a learning problem using Equation (\ref{eqn:pac}). As a consequence this learning approach is known as \emph{ Probably Approximately Correct (PAC) learning}. Evaluation of Equation  (\ref{eqn:pac}) is usually done using inequalities from \emph{concentration of measure}, see \cite{bousquet2004introduction} for the details of how the concentration of measure can be used to make statements in the form of Equation (\ref{eqn:pac}). \cite{Germain.2009} present  a  general  PAC-Bayes  theorem
from which all known Bayes risk bounds are obtained as particular cases for linear classifiers.

\subsection{Supervised Learning and Generalized Linear Models}

The Computer Science community sees  supervised learning as solution to two different problems, namely (i) Regression and (ii) Classification.  However, the Statistics community sees supervised learning as a single solution class and models it using the generalized linear models (GLM), see e.g., \cite{McCullagh.Nelder.1989}. The approach in Statistics is to model the dependent variable $y$ using the natural exponential family
$$
p(y)=\exp\{\theta y - \psi(\theta)\},
$$
for different choices of $\theta$, $p(\cdot)$ may represents a  Gaussian, Binomial, Bernoulli and Poisson distribution, see \cite{McCullagh.Nelder.1989} for detail. These distributions can model a continuous, binary or count response variables, see \cite{Das.2006,Das.Dey.2013}. Then features are modeled as the function of the conditional mean of $y$, through the link function $g(\cdot)$ as 
$$
\mathbb{E}(y|\bm{x})=g(\beta_1x_1+\cdots+\beta_px_p),
$$
where $\bm{x}=(x_1,x_2,\hdots,x_p)$ is the vector of $p$ features or covariates in the datasets. The Bayesian GLM is presented in \cite{Gelman_et_al}[Chapter 16].

\subsection{Complexity of Linear Hyper-planes: Multicollinearity and Feature Selection}\label{subsec:feature_select}

One of the popular hypothesis class, we consider is the family of linear hyper-planes. Suppose $X=[x_{ij}]_{n\times p}$ is the design matrix with $n$ samples and $p$ features.
$$
f(X)=X\beta,
$$
where $\beta=(\beta_1,\cdots,\beta_p)^T$ are the $p$ regression coefficients. The ordinary least square solutions of $\beta$ is
\begin{equation*}
\hat{\beta}_{OLS}=\underset{\beta}{\argmin} \large\{(y-X\beta)^T(y-X\beta)\large\},
\end{equation*}
can be obtained by solving the normal equations
\begin{equation}\label{eq:normal_equation}
X^TX\beta = X^Ty.
\end{equation}
If two (or more) predictors are highly correlated, that makes the system in Equation (\ref{eq:normal_equation}) \say{near singular}. It makes the solution unreliable. The \say{near singular} undesirable property of many problems are known as \textbf{\emph{multicollinearity}}, and the $L_2$ penalty on $\beta$ can fix the problem. The approach is known as the Ridge solution of the multicollinearity, see \cite{HoerlKennard1970},
\begin{equation}\label{eqn:Ridge_solution}
\hat{\beta}_{Ridge}=\underset{\beta}{\argmin} \large\{(y-X\beta)^T(y-X\beta)+ \lambda\beta^T\beta\large\}.
\end{equation}
If we compare the Ridge solution in (\ref{eqn:Ridge_solution}) with (\ref{eqn:comp_reg}), the first term 
$$
\hat{R}(\beta)=(y-X\beta)^T(y-X\beta),
$$ 
is the residual sum of squares and $C(\beta)=\beta^T\beta$ is the $L_2$ penalty on $\beta$. The objective function in Equation (\ref{eqn:Ridge_solution}) can be presented as
\begin{equation*}
p(\beta|y,X,\sigma^2) \propto \exp\large\{-\frac{1}{2\sigma^2}(y-X\beta)^T(y-X\beta)\large\}.\exp\large\{-\frac{\lambda}{2\sigma^2}\beta^T\beta\large\},
\end{equation*}
where $p(\beta|y,X,\sigma^2)$ is the posterior distribution of $\beta$, the $L_2$ penalty is proportional to the Gaussian prior distribution on $\beta$, where $(\beta|\sigma^2,\lambda)\sim N(0,\sigma^2/\lambda)$, and $y\sim N(X\beta,\sigma^2I)$ yields the likelihood function. In this case, the Ridge solution is the \emph{posterior mode} and it has a mathematically closed form solution:
$$
\hat{\beta}_{Ridge}=(X^TX+\lambda I)^{-1}X^Ty.
$$
This result implies that the Ridge learning method is the Bayesian solution which is also known as the shrinkage estimator, see e.g., \cite{friedman2001elements}. One more point we must note is that, if two predictors are highly correlated,  i.e., both the predictors inherently contained similar kind of information, then they are naturally  expected to have a similar functional relationship with $y$. Hence we need an algorithm, which keeps the predictors which are most relevant in predicting $y$ and drop the less crucial features and come up with a parsimonious model, see \cite{Tibshirani1996lasso}. Managing the complexity of the hypothesis class involves reducing the number of features in $f$ and the task is known as the \emph{feature selection}, see \cite{Tibshirani1996lasso}. In Bayesian statistics, the same task is known as the \emph{model selection}, see e.g. \cite{gelfandanddey1994}.

Hence, the learning algorithm $\mathcal{A}$ should figure out the best subset of $p$ features from $X,$ for which a performance metric like the Mean Square Error (MSE) is minimum or the adjusted-$R^2$ is maximum. One can apply the \emph{best subset selection}, see \cite{friedman2001elements} [Chapter 3], but the best model has to search through $2^{p}$ many models. So the complexity of model space makes it impossible to implement even for a dataset with $p=20$ different features. The \emph{forward-stepwise subset selection} is a greedy algorithm and the model complexity is $\mathcal{O}(p^2)$.

In a recent paper, \cite{Bertsimas.2016} show that the traditional best subset selection puzzle formulated as a mixed integer optimization (MIO) problem. With the recent advances in MIO algorithms, they demonstrate that the best subset selection obtained at much larger problem sizes than what was thought impossible in the statistics community. It is a new alternative approach compared to popular shrinkage methods.

The shrinkage methods are a popular technique to manage complexity for linear hyper-planes hypothesis class, see \cite{Tibshirani1996lasso}.  The Least Absolute Shrinkage and Selection Operator  (LASSO) can be a particularly effective technique for feature selection, see \cite{Tibshirani1996lasso}. If the values of coefficients are estimated to be zero, then effectively the solution is to drop that feature from the model. Such solutions are called \emph{sparse} solutions. The LASSO yields the desired \emph{sparse} solutions with $L_1$ penalty on $\beta$, defined as
$$
C(\beta)=\lambda \sum_{j=1}^{p}\big|\beta_j\big|.
$$
Although, the Ridge solution handles the multicollinearity issue, it, however,  fails to yield the \emph{sparse} solutions. The LASSO estimate is defined as:
\begin{equation} \label{eqn:lasso}
\hat{\beta}_{lasso} = \underset{\beta}{\argmin}\bigg\{ (y-X\beta)^T(y-X\beta) + \lambda \sum_{j=1}^{p}\big|\beta_j\big| \bigg\},
\end{equation}
where $\lambda$ is a parameter that affects the \emph{sparsity} of the solution. The $L_1$ penalty on  $\beta$ is equivalent to the Laplace or double exponential prior distribution, see e.g., \cite{Park.Casella.2008}. The LARS algorithm for LASSO solution is a popular algorithm which makes the LASSO solution highly scalable for large datasets.

Note that $\lambda$ is a parameter that must be provided to the learning algorithm $\mathcal{A}$. There are several approaches to learn $\lambda$. In one approach, $\lambda$ is learned using a grid search with $k$-fold cross-validation technique. In another approach, full Bayesian methodology elicits a prior on $\lambda$, known as the Bayesian LASSO presented in \cite{Park.Casella.2008}. The Bayesian LASSO focuses on estimating the posterior mean of  $\beta$ using the Gibbs sampler. The slow implementation of the Gibbs sampler makes the full Bayesian implementation of the LASSO  less attractive for  practitioners. On the contrary, the fast, scalable implementation of the LARS makes it very attractive with partial Bayes solution for the practitioner. 

The convex combination of the $L_1$ and $L_2$ penalty yields a new kind of penalty, known as the \emph{elastic net},
\begin{equation} \label{eqn:elastic_net}
\hat{\beta}_{EN} = \underset{\beta}{\argmin}\bigg\{ (y-X\beta)^T(y-X\beta) + \lambda \sum_{j=1}^{p}\big(\alpha\big|\beta_j\big| + (1-\alpha)\beta_j^2\big)\bigg\},
\end{equation}
where $0\leq \alpha \leq 1$, see \cite{ZouHastie2005}. Like LASSO and Ridge, we can similarly argue that the Elastic Net solution is a Bayesian solution and fully Bayesian Elastic Net implementation is also available, see \cite{LiLin.2010}. One of the advantages of the Elastic Net is that it can address the multicollinearity problem and feature selection together. The \emph{copula prior} proposed in a recent paper showed that the Ridge, LASSO, elastic net etc. are special cases of the copula prior solution, see \cite{SharmaDas2017}.

\subsection{Tree Models and Its Complexity}

Regression trees partition the input space into regions with a constant response for each region, see \cite{breiman1984classification}. The hypothesis class for trees takes the following form:
\begin{equation}\label{eqn:tree_model}
f(x) = \sum_{m=1}^{M}c_i \mathbf{I}(x \in R_m),
\end{equation}
where:
\begin{itemize}
\item $c_m$ represents the constant response for region $R_m$.
\item $\mathbf{I}(x \in R_i)$ is the indicator function that is defined as $ I(x) = \begin{cases} 1 & \text{if } x \in R_m\\
 0       & \text{otherwise}
 \end{cases}$
 \item $M$ represents the number of terminal nodes and is an important parameter for the tree hypothesis class. 
\end{itemize}
If we use the square error loss function then the optimal choice for $c_m$ is the average of the response values $y_i$ in the region $R_m$. The input space is partitioned into regions $R_1, \cdots, R_m$ using a greedy algorithm, see \cite{friedman2001elements}[Chapter 9, Section 9.2.2] for details. The number of regions ($M$), which partition the input space, is an important parameter to the algorithm. The parameter $M$ represents the height of the tree. It determines the complexity of the solution and the complexity management strategy must monitor the parameter. A strategy that works well is to partition the input space until there is a minimum (threshold) number of instances in each region. This tree is then shortened using pruning. Pruning is facilitated by minimization of a cost function which is defined as follows.
\begin{itemize}
\item Let $N_m$ be  the number of instances that belong to region $R_m$ and $c_m = \frac{1}{N_m}\sum\limits_{X_i \in R_m} y_i$,
\item Let $T_0$ represent the tree  obtained without applying the pruning by developing the tree until a minimum number of instances in each leaf node is  achieved.
\item Let $T$ be the tree that is subject to pruning. The pruning process involves collapsing nodes of $T_0$ to build an optimal tree. It has $\big|T\big|$ nodes.
\item Define mean sse as : $Q_m(T) = \frac{1}{N_m}\sum\limits_{y_i \in R_m}\big(y_i - c_m\big)^2$
\end{itemize}
We can  define the cost function that is minimized during ERM with the regression tree hypothesis class as:
\begin{equation}\label{eqn:tree_pruning}
C_{\alpha}(T) = \sum\limits_{m=1}^{\big|T\big|} N_m Q_m(T) + \alpha \big|T\big|,
\end{equation}
where $\alpha$ is a parameter that controls the complexity associated with $T$. Note that $C_{\alpha}(T)$ is the penalized sum of square of errors. As with the linear model, for each value of $\alpha$, we obtain a hypothesis $f_{\alpha}$ by applying ERM where the Equation (\ref{eqn:tree_pruning}) is minimized. Many variations of tree models are developed. One of the most popular is the \textbf{Random Forest}. The random forests is an ensemble learning of tree models, see \cite{breiman2001random}. The ensemble learning is discussed in Section \ref{sec:ensemble_learning_approach}.

\subsection{Gaussian Process Prior for Machine Learning}\label{subsec_GPprior}

The final hypothesis class we consider is the Gaussian Process priors. For this, we consider a full Bayesian approach to learning. We still use the template defined by  Equation (\ref{eqn:comp_reg}); however we now use the Bayesian approach, explained in the Equation (\ref{eqn:bayes_rule}), to pick the best model from the hypothesis class $\mathcal{F}$.

A Gaussian Process  is viewed as a prior over a space of functions. We encode our beliefs about the family of functions that are suitable for the problem by choice of a covariance function or kernel. In other words, $\mathcal{F}$ is specified by using this covariance function or kernel, see \cite{Rasmussen.2006} for further details. See \cite{duvenaud-thesis-2014} for guidelines about picking a kernel for a problem.

We consider the generic hyper-plane presentation
$$
y=f(x)+\epsilon,
$$
where $\epsilon\sim \mathcal{N}(0,\sigma^2 I)$. This means $y\sim N(f(x),\sigma^2 I)$, 
$$
f(x)=\sum_{i=1}^{\infty}\beta_k\phi_k(x),
$$
$\phi_k(x)$ is a completely known basis function and $\beta_k$'s are unknown, uncorrelated random variable from Gaussian distribution, then the Karhunen-Lo$\grave{e}$ve theorem states that $f(x)=\beta\phi$ follows the Gaussian process. Here the hyper-plane $f(x)$ is completely unknown and we assume that $f$ is a random realization from the Gaussian process. Therefore the resulting model is known as the Gaussian process prior model, see \cite{Rasmussen.2006}. The corresponding model for data is
$$
y \sim \mathcal{N}_n(\phi(x)\beta_0, K(x,x')+\sigma^2I_n),
$$
where $E(f)=E(\phi(x)\beta)=\phi(x)E(\beta)=\phi(x)\beta_0$ is the prior mean function, such that $\beta\sim \mathcal{N}(\beta_0,\sigma^2I)$, and $K(x,x')$ is the covariance kernel function of $f$. The estimated value of $y$ for a given $x_0$ is the mean (expected) value of the functions sampled from from the posterior at that value of  $x_0$. Suppose $\mu(x) = \phi(x)\beta_0$, then expected value of the estimate at a given $x_0$ is given by
\begin{eqnarray}\label{GP_Prior_solution}
\hat{f}(x_0)&=& E(f(x_0|x,y)) \nonumber\\
     &=& \mu(x_0)+K(x_0,x)\large[K(x,x')+\sigma^2 I_n\large]^{-1}(y-\mu(x)). \nonumber
\end{eqnarray}
If we choose the prior $\beta \sim \mathcal{N}(0,\sigma^2 I)$, then the above equation can be expressed as
\begin{eqnarray}
\hat{f}(x_0)&=&K(x_0,x)\large[K(x,x')+\sigma^2 I_n\large]^{-1}y. \nonumber
\end{eqnarray}
One very strong point in favor of the Gaussian process prior model is that the $\hat{f}(x)$ approximates $f(x)$ well, i.e.,
$$
P\big(\sup_x|\hat{f}(x)-f(x)|<\epsilon\big)>0~~\forall \epsilon>0,
$$
see \cite{Ghoshal.Book.2017} for further details.  However, the solution  $\hat{f}$ involves the inversion of the covariance matrix of order $n$. The time complexity of the matrix inversion is $\mathcal{O}(n^3)$ and the space complexity is $\mathcal{O}(n^2)$. This limits the applicability of the Gaussian process prior model to small and moderate size dataset. Recently, a fast bagging algorithm has been developed to  implement the Gaussian process prior regression for massively large dataset and makes the GP prior highly scalable, see \cite{das2015fast}.

\subsection{Is P-Value Missing from Machine Learning Literature?}

In the dictionary presented in Table \ref{tbl_dictionary}, we see corresponding to the ``hypothesis Testing," no concept is available in the Computer Science. Though computer scientists are very much aware of the theory of the hypothesis testing in Statistics and they often use it in practice;  it is rarely discussed or mentioned in the mainstream machine learning literature. The reasons are the following:
\begin{itemize}
\item There is a lot of development in SML by technology companies, see e.g., \cite{shinal_google_2017}. Therefore the SML focus more on the prediction of the target variable, rather than finding association between target variable and features, and hypothesis testing whether a particular feature influence influence the target variable or not.
\item Even if, one is making inference; typically  SML deals with massive datasets, which requires handling of thousands of hypothesis. It results in the multiple testing problems.
\item A focus on statistically significant, novel and affirmative results, based on the \say{P-value} method leads to bias, known as \say{P-hacking.}  When researchers keep collecting data until  analyses yield statistically significant results, the studies become prone to \say{P-hacking.} Results found from \say{P-hacking} are generally not reproducible. The testing of the hypothesis, if not correctly conducted, unknowingly it might lead the scientists to P-hacking, a situation all scientist must avoid, see \cite{P_hacking_2015} for detail.
\item The SML considers each hypothesis as a model and Bayesian model selection strategy is a much more comprehensive strategy for large data.
\end{itemize}


\section{Challenges to Machine Learning for  Big Data}\label{sec:mlbd}

Organizations that have made investments in big data desire to exploit their data assets to gain operational efficiencies or business advantage. From a learning theory standpoint, it would seem that having  data to perform learning is a good thing. However, \cite{mckinsey_how_nodate} presents an organizational perspective of challenges and hurdles in applying machine learning to big datasets. Computer scientists use time and space complexity to characterize the running time (processing) and storage required by algorithms. The `storage required' and `run-time' characterized as functions of the size of the dataset, i.e., in terms of $n$ and $p$. Ideas from the order of growth of functions are used to characterize the asymptotic behavior of the running time and storage associated with an algorithm, see e.g., \cite{cormen2001introduction}[Chapter 3]. The upper bound of a function $f(n)$ used to characterize the time complexity, or space complexity is described using the big-$\mathcal{O}$ notation. The upper bound of $f(n)$ in the big-$\mathcal{O}$ notation  described in Equation (\ref{eqn:big_oh}):
\begin{eqnarray}\label{eqn:big_oh}
O(g(n)) = \{f(n): &\exists& ~ \mathrm{positive\ constants}\  c, n_0, \\
&& \ \mathrm{ such \ that} \  0 \leq f(n) \leq c.g(n),\ \forall n \geq n_0\}.\nonumber 
\end{eqnarray}
Several commonly used machine learning algorithms are associated with polynomial space and time complexity. For example, kernel methods are associated with $\mathcal{O}(n^3)$ time complexity and $\mathcal{O}(n^2)$ space complexity, see e.g.,  \cite{shawe2004kernel}. Therefore when $n$ is large (say one million samples), time and space complexity is beyond the available computational resources. In high dimensional datasets, the dimensionality of the problem creates the computational bottleneck. For example, principal component analysis (PCA) on a high dimensional dataset is associated with time complexity $ \mathcal{O}(np^2 + p^3)$, see e.g., \cite{tipping1999mixtures}. Large values of $p$  like those encountered in genomic  or text mining datasets impose computational challenges to dimension reduction techniques. Overcoming these challenges requires ideas from both Computer Science and Statistics. The challenges in application of data mining to big data is described in \cite{wu2014data}.  Recently, \cite{l2017machine} survey the challenges in applying machine learning to big datasets. The paper provides the challenges in the context of the characteristics of big datasets described in Section \ref{sec:bd}. We describe three other approaches to machine learning that are not mentioned in \cite{l2017machine}. In this paper, we motivate the approaches to machine learning on big datasets using ideas from statistical learning theory discussed in Section \ref{sec:slt}. We provide the particular examples of these perspectives and a description of how they mitigate the computational challenges.

\subsection{Approaches to Machine Learning for Big Data} \label{sec:app_mlbd}
We adopt the terminology used in \cite{l2017machine} to discuss the approaches. We provide a brief description of each approach to provide a context for discussion. 

\subsubsection{\textbf{Online Learning}}

There are two ways machine learning algorithms interact with the data. One is the \emph{batch learning}, and the other is the \emph{online learning}. In \emph{batch learning}, the algorithm processes the entire dataset. In \emph{online learning}, the algorithm receives data sequentially. The model trained on an initial dataset that is used to predict an incoming stream of observations, one sample at a time. This idea is similar to \textbf{\emph{sequential analysis}} in Statistics, see e.g., \cite{Berger2017}. After the prediction of the incoming observation, the model updates the performance of the predicted observation (the feedback) into consideration, see e.g., \cite{shalev2007online} and \cite{shalev2008tutorial} for algorithmic approaches and theory. \cite{Opper.1999} discussed  online learning from a Bayesian inference point of view. Bayesian online learning consists of prediction steps and update steps for learning of unknown parameter $\theta$. The prior predictive mean
$$
E(\theta|D_{t})=\int \theta \, p(\theta|D_t)d\theta,
$$
where $p(\theta|D_t)$ is the prior predictive distribution 
$$
p(\theta|D_{t})=\frac{p(D_t|\theta)p(\theta)}{\int_{\theta\in \Theta}p(D_t|\theta)p(\theta)d\theta},
$$ 
where $D_t=\{y_t,\cdots,y_1\}$. The learning of $\theta$ updates itself when new sample $y_{t+1}$ arrives with the posterior mean as
$$
E(\theta|D_{t+1})=\int\theta p(\theta|D_{t+1})d\theta,
$$
where the posterior density is
$$
p(\theta|D_{t+1})=\frac{p(y_{t+1}|\theta)p(\theta|D_t)}{\int_{\theta\in \Theta}p(y_{t+1}|\theta)p(\theta|D_t)d\theta}.
$$
Applications may consider online learning for one of the following reasons:
\begin{enumerate}
\item There is too much data to keep in memory.
\item The nature of the application may be such that the hypothesis class is continuously evolving. This is a characteristic of many application domains like  advertisement, online user experience personalization, etc. and is called \emph{concept drift}. 
\end{enumerate}
On-line applications  are associated with higher deployment complexity than batch models since these require a model update with every request processed. The choice of online versus batch learning for a particular task depends on the questions and organizational objectives associated with the problem. Batch learning requires the learning algorithm to process all of the data in the dataset. It creates computational challenges. We need to apply a suitable choice from the other approaches described below to overcome these challenges. Many machine learning algorithms have online implementations. \cite{laskov2006incremental} describe an application of support vector machines in an online manner.
\cite{Das.Dey.2013} present the online learning of Generalized Linear Models. \cite{sambasivan_statistical_2017} describe an online approach with Gaussian Processes prior models.

\subsubsection{\textbf{Posterior Mode with Large Scale Optimization}}\label{subsub_post_mode_large_opt}

As we discussed in Section \ref{subsec_ERM} and \ref{subsec:feature_select}; finding the posterior mode as the statistical machine learning solution,  optimization methods are used extensively. Optimization problems are represented as below:
\begin{equation}
\begin{aligned}
& \underset{\theta}{\text{minimize}}
& & f_0(\bm{\theta}) \\
& \text{subject to}
& & f_i(\bm{\theta}) \leq b_i, \; i = 1, \cdots, m.
\end{aligned}
\label{eqn:canonical_form_opt}
\end{equation}
Here $\bm{\theta}=(\theta_1, \theta_2,\cdots, \theta_p)$ are the parameters being estimated and $f_0 : \mathbf{R}^p \mapsto \mathbf{R}$ is the function to be optimized and is called the \emph{objective function}, see \cite{boyd2004convex}. The empirical risk (in Equation (\ref{eqn:emp_risk})) is the objective function that needs to be minimized and $f_i : \mathbf{R}^p \mapsto \mathbf{R},\ i=1,2,\cdots, m$ are called the constraints. The constants $b_1, b_2, \hdots, b_m$ are called the limits for the constraints. The constraints correspond to the penalty or cost function $C(f)$ in penalization method in Section \ref{subsec_penalization_method}.  \\

An optimization problem is \emph{convex} if the objective function and the constraints are convex. This means that these functions satisfy the conditions below:
\begin{eqnarray}
\quad f_i(\sum_{j=1}^p\alpha_j \theta_j ) \leq \sum_{j=1}^{p}\alpha_j f_i(\theta_j) , ~~i = 0,1,\cdots, m;~~ \alpha_i \in \mathbf{R},~~ \sum_{j=1}^p\alpha_j = 1.
\label{eqn:convex_function}
\end{eqnarray}
When an optimization problem is convex, then it has a unique minimum. The \emph{empirical risk minimization} (ERM) principle discussed in Section \ref{sec:slt} is often implemented using optimization. The risk function corresponds to the loss function that is appropriate for the problem. Many machine learning problems are convex optimization problems and numerical optimization techniques are used in these problems, see e.g.,  \cite{wright1999numerical}. As with hypothesis classes for learning theory, there are many algorithms for numerical optimization. The appropriate choice depends on problem characteristics. As with hypothesis classes, no optimization algorithm is optimal for all problems. This result is called the \emph{\say{No Free Lunch Theorem for Optimization}}, see e.g., \cite{wolpert1997no}. Two numerical optimization approaches are predominantly used to perform ERM on large datasets, namely (i) \emph{stochastic} and (ii) \emph{batch} approaches. The details of both these techniques as well as a comprehensive review of optimization methods in machine learning are provided in \cite{Bottou.2017}. Large scale optimization techniques are used to solve both \say{big n} and \say{big p} problems discussed in Section \ref{sec:bd}. An example of application to a \say{big n} problem is in  \cite{zhang2004solving}.  An example of application to a \say{big p} problem is available in  \cite{lu2012convex}.

Many problems solved using machine learning are non-convex. The most prominent machine learning technique associated with non-convex optimization is deep learning (discussed later in this section). An overview of the techniques for non-convex optimization along with applications is presented in \cite{jain2017non}.

\subsubsection{\textbf{Bayesian Inference}}

Bayesian inference methods require  computation of posterior probabilities that are often analytically intractable when the datasets are large. \cite{zhu2017big} provide a discussion of methods used to apply Bayesian methods to big datasets. Bayesian inference has been applied to both \say{big n} and \say{big p} problems. Two methods predominantly used are: 
\begin{itemize}
\item Markov Chain Monte Carlo Methods (MCMC): These methods use repeated random sampling to approximate the posterior distribution. \cite{andrieu2003introduction} present the conceptual building blocks of the MCMC technique and provide an overview of various MCMC algorithms used in machine learning.
\item Variational Inference: This is a method to approximate posterior probabilities using an optimization based machine learning approach. The objective is to approximate the posterior density and to do so we consider a family of candidate densities. The approximation is measured using the Kullback-Liebler divergence between the target and candidate posterior density. \cite{blei2017variational} provide a discussion of Variational Inference and its applications to solving big data problems.
\end{itemize}
A significant development in the application of Bayesian approaches to machine learning is the development of \emph{Bayesian Non-Parametric} methods. Bayesian Non-Parametric methods adapt the complexity associated with the hypothesis class based on the data observed. In a parametric approach, the complexity of the hypothesis class is specified a priori and does not adapt to the data received by the learning algorithm. Using a parametric hypothesis class requires the use of other methods to manage the level of complexity of the hypothesis class using the data, see e.g., \cite{nowak_2018_slt_l3}[Section 2.2]. \cite{gershman2012tutorial} provide an introductory overview of Bayesian Non-Parametric methods. 

\subsubsection{\textbf{Local Learning}}
Complex hypothesis classes are often associated with high computational complexity. For example, as discussed in Section \ref{subsec_GPprior}, Gaussian Processes prior or kernel methods are associated with $\mathcal{O}(n^3)$ time complexity. Therefore applying a complex hypothesis class to a \say{big n} dataset makes learning intractable. A natural question to consider with \say{big n} datasets is: \say{would a divide and conquer approach  make the computation tractable ?}. It does. Different strategies can be used to partition the dataset, and we could apply the learning procedure discussed in Section \ref{sec:slt} on the partitions. We can consider a complex hypothesis for the partitions because the size of the partition is much smaller than the size of the dataset and computational complexity is not a limiting factor in the choice of the hypothesis class. \cite{park2010hierarchical} and \cite{tresp2000} provide two approaches for applying Gaussian Processes on big data that use clustering to create the partitions. A decision tree algorithm, for example the Classification and Regression Tree algorithm (CART), can be used to create the partitions, see e.g., \cite{breiman1984classification} and  \cite{sambasivan_bdrtbs_2017}.

\subsubsection{\textbf{Ensemble Learning}}

\label{sec:ensemble_learning_approach}
In \emph{ensemble learning}, we consider several learners, as discussed in Section \ref{sec:slt} for a particular learning task. The ensemble model combines the predictions from each of the learners to produce the final prediction. There are a couple of ways to achieve this - \emph{bagging} and \emph{boosting}. Bootstrap aggregating, also called bagging,  achieves good performance by reducing the estimation error discussed in Section \ref{sec:slt}, see \cite{breiman1996bagging}. Boosting achieves good performance by reducing the approximation error discussed in Section \ref{sec:slt}. In practice, trees are one of the most used hypothesis class to build ensembles. \textbf{Random forests} is an ensemble algorithm that is based on bagging tree models, see \cite{breiman2001random}. Extreme gradient boosted trees, aka., \textbf{xgboost} is an ensemble algorithm that is based on boosting tree models, see \cite{chen2016xgboost}. Both \textbf{random forests} and \textbf{xgboost} are scalable and can be used with big datasets. Ensemble techniques are used with \say{big n} datasets. The bagging and boosting framework are generic. 

\cite{Chipman.EdGeorge.2006} presented a Bayesian ensemble learning method, where each tree is
constrained by a prior to be a learner. Fitting and inference were accomplished
using  the MCMC algorithm. \cite{Quadrianto.2015} presented the Bayesian random forest, by random sampling many trees from a prior distribution, and perform weighted ensamble of predictive probabilities. Hypothesis classes other than trees can be used with these techniques. For example, \cite{das2015fast} apply bagging using Gaussian Processes prior regression on big datasets. Model development in ensemble methods, such as \textbf{xgboost}, is performed on a representative sample from the data, see e.g., \cite{chen2016xgboost}. This makes these methods very scalable.

\subsubsection{\textbf{Deep Learning}}
In Section \ref{sec:slt} we have described that the goal of the statistical machine learning (SML) is to learn a function that can predict labels accurately. The performance of the function is impacted by the features. If we have good features, then we can achieve good performance with enough data and an appropriate hypothesis class. Deep learning exploits this idea. In deep learning terminology these features are called \emph{representation}, see \cite{Goodfellow-et-al-2016}[Chapter 1]. For many SML problems generating highly discriminative features from the observed set of features is very difficult. In deep learning, supervised learning algorithms learn both good features and the function that maps the feature space to the output space. Deep learning also provides algorithms for learning good features or representations. This type of learning is unsupervised and is called representation learning, see \textbf{auto-encoders} in  \cite{Goodfellow-et-al-2016}. Deep learning algorithms also use optimization for empirical risk minimization. \\

Back-propagation is a technique used to calculate the gradients of the risk function in the optimization procedure used for deep learning. Deep learning is computationally intensive. Interestingly, the hardware architecture for the video graphics cards used in computers (the GPU) can perform the computation associated with back-propagation very efficiently. It led to new hardware development for deep learning. For a discussion of optimization techniques in deep learning see \cite{Goodfellow-et-al-2016} and \cite{Bottou.2017}. Deep learning can be applied to both \say{big n} and \say{big p} problems, see \cite{Goodfellow-et-al-2016}.   Practical suggestions to develop effective deep learning applications are presented in \cite{goodfellow_practical_video}. The uncertainty estimation for Bayesian deep learning for computer vision is presented by \cite{Kendall.Gal.2017}. To the best of our knowledge Bayesian deep learning is an emerging field and many research issues are yet to be addressed.
 
\subsubsection{\textbf{Transfer Learning}}
When we have limited labeled data, the \emph{transfer learning} learns from different domain which have plenty of labeled data, and transfers the relevant knowledge to the target domain with an optimal Bayesian update, see e.g., \cite{Karbalayghareh.2018},  \cite{l2017machine}, and  \cite{pratt1993discriminability}.  \emph{Lack of data for a problem} is what motivates transfer learning. This idea is similar to the idea of \textbf{prior elicitation} in Bayesian statistics, using either expert opinion, or older data or based on the published journal article, see e.g., \cite{Das.Xia.Banks.2012}. A lot of the reported successes in this area are associated with deep learning. \cite{yosinski2014transferable} discuss the transferability of features in deep learning. The system whose model utilizes is called the \emph{source} system and the system for which limited data is available is called the \emph{target} system. While the target system may not have sufficient data, the source system may train on big data. 

\subsubsection{\textbf{Life Long Learning}}
We concluded Section~\ref{sec:slt} by saying that achieving good performance with a learning algorithm requires incorporating prior knowledge of the problem and sufficient data. Big datasets may provide the latter. However, we need a framework to combine the experience with the problem (prior information) into the learning algorithm. The Bayesian framework becomes handy here. Lifelong learning considers systems that can learn many tasks over a lifetime from one or more domains. They efficiently and effectively retain the knowledge they learned and use that knowledge to more efficiently and effectively learn new tasks, see e.g.,  \cite{silver2013lifelong}, \cite{zhiyuan_chen_kdd_video} and \cite{chen2016lifelong} for an overview of lifelong learning. The PAC-Bayesian bound for lifelong learning is presented by \cite{Pentina.2014}.

\subsection{Hyper-Parameter Tuning with Bayesian Optimization}
Approaches to statistical machine learning (SML) like ensemble methods and deep learning require us to set parameters that are not part of the empirical risk minimization (ERM) procedure associated with the learning task, see Section \ref{sec:slt}. These parameters are called \emph{hyper-parameters}. Examples of hyper-parameters are the following:
\begin{itemize}
\item The number of trees to use in a random forest model.
\item The fraction of the data to use for sampling in an ensemble tree method - either xgboost or random forests.
\end{itemize}
These hyper-parameters are not determined during ERM but  have a significant impact on the performance of the learning task. The conventional way of choosing these parameters is to use techniques like grid search or random search. Both these techniques pick values in the hyper-parameter space and then evaluate the performance for each set of values of the hyper-parameters. The hyper-parameter settings associated with the best performance is then picked. Grid search picks the hyper-parameter settings over a grid of values for hyper-parameter space whereas random search picks hyper-parameter values based on distribution specifications we specify for the hyper-parameter. \cite{bergstra2012random} provide an example of research claims that random search performs better than grid search for deep-learning. SML solutions associated with complex hypothesis classes have many hyper-parameters. Determining good values for them is critical to achieving good performance. \cite{goodfellow_practical_video} suggests that it is prudent to develop solutions, that the practitioner knows how to tune well rather than using approaches where good hyper-parameter settings for a particular application is unknown.

A recent development in this area is the use of \textbf{\emph{Bayesian Optimization}} to pick good hyper-parameter values, see e.g.,  \cite{snoek2012practical}. The approach used is to model hyper-parameter selection as an optimization problem. A characteristic of this problem is that there is no explicit form of the objective function. A Gaussian Process model is used as a proxy for the objective function. The theme for determining good hyper-parameter values is to balance exploration (trying new hyper-parameter values) versus exploitation (using hyper-parameter values for which performance is known). Experiments to tune hyper-parameters are computationally intensive in big datasets. Bayesian optimization mitigates this problem by performing experimentation in a principled manner.

\section{Some Practical Issues in Machine Learning}\label{sec:spiml}
There is a lot of development in Statistical Machine Learning (SML) by technology companies, see \cite{shinal_google_2017}. While the sentiment around SML as technology is upbeat, it is hard to escape the hype around it, see \cite{williams_ai_nodate}. \cite{larose2005discovering}[Chapter 1] provides an excellent discussion of the fallacies and the realities in this area. It gives an indication of the dissonance between perception and reality that practitioners face in applying machine learning and data mining. At this point, SML as a technology requires several human subject matter experts to implement it successfully. For data mining projects, process models to apply data mining have been developed, see \cite{foroughi2018data} . As discussed in \cite{larose2005discovering}[Chapter 1], it is quite easy to perform SML poorly. Applying a process to SML projects can ensure avoidance of many pitfalls. A typical sequence of phases in a project involves the following:
\begin{enumerate}
\item \textbf{Problem Definition and planning}: The problem that the organization or business or academic research group wants to solve using its data must be specified. One should finalize whether to use GP prior models or Decision tree or K-means clustering at this planning stage. Whether the goal is the prediction or statistical inference, should be part of the planning stage. 
\item \textbf{Data Processing}: The data relevant to the problem must be identified in the organizations' data assets. Suitable processing must be applied to transform it into the form that can be used for machine learning. 
\item  \textbf{Modeling}: The nature of the problem dictates the machine learning tasks that are required or considered for the project.
\item \textbf{Evaluation}: One must evaluate the results from the machine learning tasks in the context of the business or organizational problem.  Though `square error' or `0-1 loss' is widespread; it is better to define a metric meaningful to the problem at hand.
\item \textbf{Deployment}: If the model is deemed to be useful in the evaluation phase, it gets deployed in the live environment for public/general use. 
\end{enumerate}
A comprehensive discussion of the individual phases can be seen in  \cite{larose2005discovering} and \cite{kuhn2013applied}. Data collected by organizations is seldom in a form that is suitable for SML tasks. A significant amount of time and effort for machine learning projects is often required for data preparation. Common issues include identifying redundant data, handling missing data values and handling data that are not in a consistent state because they were gathered before transactions or organizational processes had been completed, see, e.g.  \cite{larose2005discovering} [Chapter 2] for a discussion of data pre-processing.

The machine learning algorithms we consider for a particular project or problem may dictate some transformations to the data. Scaling, centering and deriving new features are some of the commonly performed tasks during the preprocessing phase, see \cite{larose2005discovering} and \cite{kuhn2013applied}. The term \emph{feature engineering} is used to describe the process of encoding features in a manner suitable for a machine learning problem. Data collected for the machine learning typically have many extra attributes. Redundant attributes often are  associated with many adverse effects, such as multicollinearity and over fitting.

The number of samples required for a learning task, with specified accuracy, dramatically increases  as we increase the number of attributes. It is called the \emph{curse of dimensionality}, see e.g., \cite{friedman2001elements}[Chapter 2, Section 2.5] for detail. We have finite computational resources to perform the learning task. When redundant attributes are present, we waste some of the computational resources on determining model components that are not useful. These resources could have been better utilized if they were applied in estimating a more complex hypothesis class with a lesser number of variables. Redundant variables can cause over-fitting.  Redundant variables increase the computation required to determine the best function in a hypothesis class. Finally, redundant attributes make the machine learning solution less interpretable. For all these purposes, it is essential to pick an optimal subset of the attributes for the machine learning task. Hence the problem of the best subset selection is significant. 

Statistical methods likes Principal Component Analysis (PCA) and Factor Analysis (FA) are useful to reduce the number of attributes we consider for the learning task, see e.g., \cite{larose2006data}. While feature selection methods determine a subset of attributes in the original dataset, PCA and FA apply transformations to determine a set of uncorrelated variables that are called \emph{extracted features}. The entire original set of attributes considered in this transformation. PCA and FA are examples of methods that are called \emph{feature extraction} methods. The feature selection methods like LASSO, elastic net etc. try to come up with subset of features, see e.g., \cite{Tibshirani1996lasso} and \cite{ZouHastie2005}.

Many practical suggestions for developing statistical machine learning (SML) solutions are provided in \cite{Goodfellow-et-al-2016, goodfellow_practical_video}. Though these suggestion are made in the context of deep learning, a lot of this advice applies to supervised learning tasks in general.  It guides the choice of the hypothesis classes we want to consider for the project. The level of noise in the data is another consideration. If the data are very noisy, then the return on investment (ROI) to develop sophisticated models may not pay off. Noise removal using techniques such as those discussed in \cite{xiong2006enhancing} may help. In SML projects, we often have a collection of models we may consider for a particular learning task. The problem of selecting the best model in this collection is called \emph{model selection} and is typically based on some theoretical metric (commonly used choices based on information theory). Complex hypothesis classes tend to be associated with many parameters. Accurate tuning of these parameters is critical to achieving good performance with the model. It may be prudent to consider a small set of sophisticated hypothesis classes that we know how to tune well, see \cite{Goodfellow-et-al-2016} and \cite{goodfellow_practical_video}. Because the parameter space for sophisticated hypothesis classes is large, it is often challenging to identify a high performing set of parameter values for unfamiliar hypothesis classes in a reasonable time. Therefore it may be expedient to consider hypothesis classes that have the complexity to address the needs of the problem and for which good hyper-parameter settings are known.

The analysis of the training and test errors can yield information that we can use to refine the hypothesis classes that we are considering for the learning task. A hypothesis class \emph{under fits} when it does not have sufficient complexity to model the variations that manifest in the data. The under fit is characterized by high test and train error. Analysis of the training and test errors can reveal if the hypothesis class we are considering for a problem under fits, over fits or is a good fit.

\section{Application}\label{sec:application}

In this section we present three applications with three different size datasets. The code used to obtain the results in  this section can be downloaded from the website: \url{https://github.com/cmimlg/SMLReview}.

\subsection{Predicting House Prices - A Regression Task with Small Datasets}

In this example we consider a real-life non-spatial regression task. The Boston housing dataset provides house prices in the Boston region, see \cite{Lichman}. The dataset contains information about census tracts from the 1970 census. Each tracts includes the median house price value. For illustration, consider a regression task where the objective is to predict the value of median house price when we know other information about the census tract. A description of the attributes of the dataset is provided in Table \ref{tab:dd_bh}.

\begin{table}[ht]
\centering
\begin{tabular}{|c|c|c|}
  \hline
 & Attribute & Description \\ 
  \hline
1 & CRIM & per capita crime rate by town \\ 
  2 & ZN & proportion of residential land zoned for lots over 25,000 sq.ft \\ 
  3 & INDUS & proportion of non-retail business acres per town \\ 
  4 & CHAS & Charles River dummy variable (= 1 if tract is adjacent to river; 0 otherwise) \\ 
  5 & NOX & nitric oxides concentration (parts per 10 million) \\ 
  6 & RM & average number of rooms per dwelling \\ 
  7 & AGE & proportion of owner-occupied units built prior to 1940 \\ 
  8 & DIS & weighted distances to five Boston employment centers \\ 
  9 & RAD & index of accessibility to radial highways \\ 
  10 & TAX & full-value property-tax rate per USD 10,000 \\ 
  11 & PTRATIO & pupil-teacher ratio by town \\ 
  12 & B & Proportion of Black Minority \\ 
  13 & LSTAT & percentage of lower status of the population \\ 
  14 & MEDV & median value of owner-occupied homes in USD 1000's \\ 
   \hline
\end{tabular}
\caption{Description of the Boston Housing Dataset}
\label{tab:dd_bh}
\end{table}

The quantity we want to predict is the median house value (attribute \emph{MEDV}) for a census tract. The other attributes represent the input. The notation used in the discussion that follows is consistent with that used in Section \ref{sec:slt}.
We  consider three hypothesis classes to learn a solution for the regression problem. There are 505 samples in the data set. We  use $70 \%$ of the data for training and  $30 \%$ of the data for testing. We  use the root mean square error as a metric to evaluate the performance of the model produced by the hypothesis class. 

First, we consider the family of linear hyper-planes. The Shrinkage Methods, as discussed in the previous section, are implemented to manage the complexity for linear hyper-planes hypothesis class.  Next we  consider the regression trees, see e.g., \cite{breiman1984classification}. We try a range of $\alpha$ values in (\ref{eqn:tree_pruning}) and pick the solution that has the lowest penalized sum of squared errors. For each $\alpha$, Equation (\ref{eqn:tree_pruning}), i.e., $C_{\alpha}(T)$ provides the optimal tree size. This is illustrated in Table \ref{tab:tree_pruning_cp}. The \texttt{rpart} package \cite{rpart_pkg} is the regression tree implementation that we use here. A review of Table \ref{tab:tree_pruning_cp} shows that a tree with $6$ splits produces the best result.

\begin{table}[H]
\centering
\begin{tabular}{|c|c|c|}
  \hline
$\alpha$ & Num Splits $|T|$ & $C_{\alpha}(T)$ \\ 
  \hline
0.51 & 0 & 1.01 \\ 
  0.17 & 1 & 0.52 \\ 
  0.06 & 2 & 0.36 \\ 
  0.04 & 3 & 0.30 \\ 
  0.03 & 4 & 0.29 \\ 
  0.01 & 5 & 0.27 \\ 
  0.01 & 6 & 0.25 \\ 
  0.01 & 7 & 0.26 \\ 
   \hline
\end{tabular}
\caption{Selection of $\alpha$ for Regression Tree, based on $C_{\alpha}(T)$. Lower the $C_{\alpha}(T)$ better it is. The best choice corresponds to $|T|=6$ and $\alpha=0.01$}
\label{tab:tree_pruning_cp}
\end{table}

\begin{figure}[H]
\minipage{0.32\textwidth}
  \includegraphics[width=\linewidth]{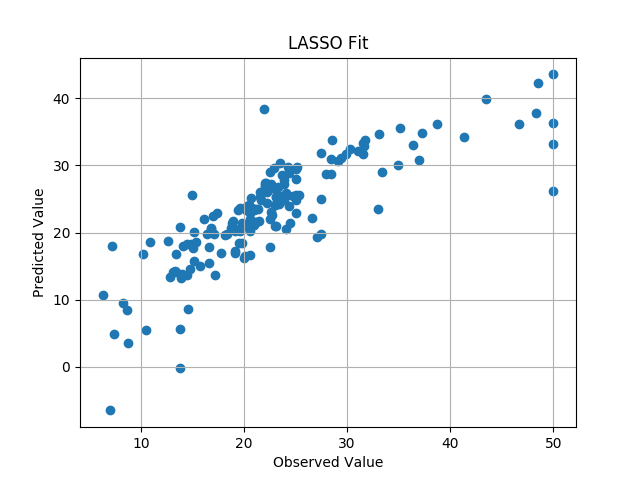}
  \caption{LASSO Fit}\label{fig:LASSO_fit}
\endminipage\hfill
\minipage{0.32\textwidth}
  \includegraphics[width=\linewidth]{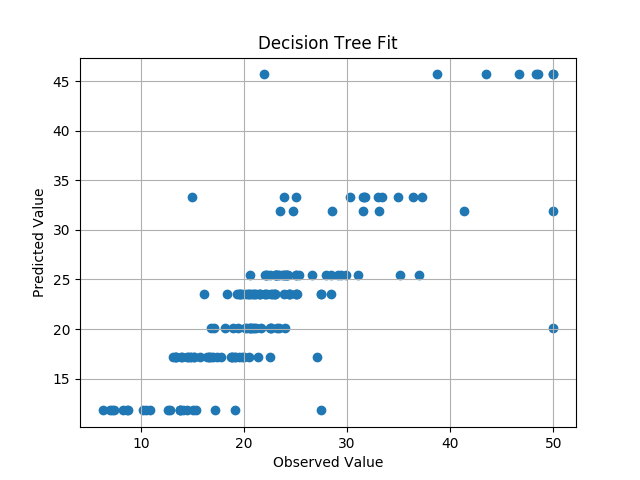}
  \caption{Regression  Tree Fit}\label{fig:DTR_fit}
\endminipage\hfill
\minipage{0.32\textwidth}%
  \includegraphics[width=\linewidth]{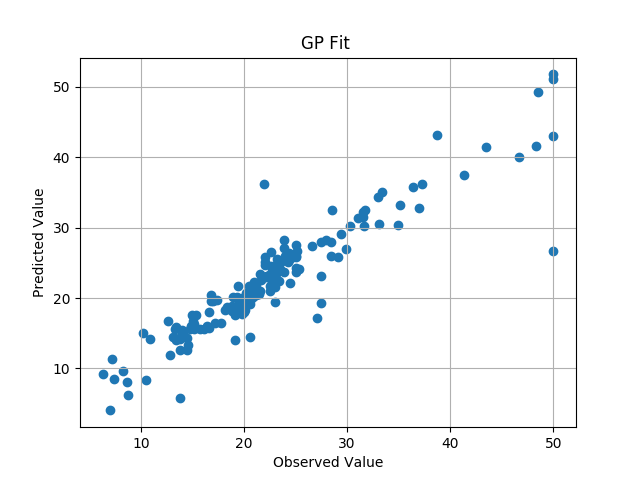}
  \caption{GP Fit}\label{fig:GP_Fit}
\endminipage
\end{figure}

Finally, we consider the Gaussian process prior models. The kernel used for this problem is a sum of a linear kernel and squared exponential kernel. Now we discuss the details of the hypothesis classes and perform learning (or estimation) using the  hypothesis classes. In Table \ref{tab:mse_bh}, we present the RMSE in the test set, for each of the hypothesis classes. The actual observed and prediction of $y$ in the test dataset from each of the models is shown in  Figures \ref{fig:LASSO_fit} - \ref{fig:GP_Fit}.
\begin{table}[H]
\centering
\begin{tabular}{|c|c|}
  \hline
  Hypothesis Class  & RMSE\\ 
  \hline
	Gaussian Process & 3.21\\ 
  	LASSO & 4.97\\ 
  	Regression Tree & 5.06\\ 
   \hline
\end{tabular}
\caption{Root Mean Square Error (RMSE) in Test Data Set.}
\label{tab:mse_bh}
\end{table}

A review of  Table \ref{tab:mse_bh} shows that the Gaussian Process hypothesis class provides the best results. An analysis of  Figure \ref{fig:LASSO_fit} - \ref{fig:GP_Fit} shows that tracts with low and high median house values are particularly challenging. The Gaussian Process hypothesis class performs better than the other hypothesis class in this respect. 

In Section \ref{sec:slt} we discussed the notion of the minimum risk associated with the learning task, $R^*$ and the associated function $f^*$. For regression problems:
\begin{equation}
f^* = E[y|X],
\end{equation}
and is called the \emph{regression function}. The minimum risk associated with the learning task, i.e., $R^*$ associated with $f^*$. See \cite{castro_2d170_notes}[Chapter 2, Section 2.2] for a proof. The function $f^*$ represents the best that the learning algorithm $\mathcal{A}$ can achieve with any hypothesis class. See \cite{van1990estimating} for a discussion of the convergence of estimators developed using techniques such as algorithm $\mathcal{A}$ to the regression function. The values in the MSE column of  Table \ref{tab:mse_bh} represent the empirical risk on the test set. A natural question to consider is how this empirical risk would vary if we consider all possible test sets. This represents the estimation error (see discussion following Equation (\ref{eqn:excess_risk_decomp}). The Probably Approximately Correct (PAC) method provides a framework to evaluate the question discussed above. For the details of how this can be applied to regression tasks that use the mean square error to measure risk, see \cite{castro_ELEN6887_notes}. Note that the discussion in \cite{castro_ELEN6887_notes} requires that the label space $\mathcal{Y}$ is bounded. It is a reasonable assumption for this example. We can assume that house prices fall within a particular range. This example provides a sketch of how statistical learning can be applied to perform learning on a real problem.

\subsection{Classification of Forest Cover Types  with Moderate Size Dataset}

To provide a concrete example of the local learning approach, let us consider the forest cover dataset, from \cite{ForestCoverData}, that is discussed in  \cite{sambasivan_bdrtbs_2017}. The data represents forest cover information from four wilderness areas in the Roosevelt National Forest of northern Colorado in the United States. Each observation in the dataset corresponds to cartographic information about a square cell ($ 30\ meters \times 30\ meters$). The cartographic information provided includes data like the elevation, slope, soil type etc.. The forest cover type represents the tree cover in that cell, for example, spruce-fir, ponderosa pine etc.. The learning task is to predict the tree cover type associated with a set of cartographic data. The dataset has over half of a million samples. To implement a local learning approach, a CART decision tree is first applied to the dataset to partition the data into relatively homogeneous segments. The notion of homogeneity for this particular example pertains to the forest cover type associated with the segment. The number of partitions, or equivalently, the height associated with the CART tree is an important parameter.  \cite{sambasivan_bdrtbs_2017} provide guidelines and describe simple experiments that can be used to estimate this parameter. The second step of the local learning approach discussed in  \cite{sambasivan_bdrtbs_2017} develops a pool of classifiers for each partition. The cross-validation error for each classifier is noted and the best performing classifier is chosen as the classifier for that partition. For the forest cover dataset, it turns out that the nearest neighbor classifier performs best in most partitions. This suggests that the intuition that trees of the same type grow close to each other can be exploited to develop a good classifier for this dataset. If a single nearest neighbor classifier is developed for the entire dataset, then this would involve the computation of a very large distance matrix to determine the nearest neighbors of each test point. When we use a divide and conquer strategy, the distance matrix computation is limited to each partition. This is computationally a much easier task. For this dataset, the local learning strategy performed better than the ensemble learning strategy discussed in the next section. The results of applying an ensemble learning strategy (xgboost, see e.g. \cite{chen2016xgboost}) and a local learning strategy is summarized in Table \ref{tab:local_lrn}
\begin{table}[H] 
	\centering
	\begin{tabular}{|c|c|c|}
		\hline
		Metric & Local Learning & Ensemble Learning \\ 
		\hline
		 RMSE & 0.916  & 0.957 \\
         AUC & 0.957 & 0.923 \\
		\hline	
	\end{tabular}
\caption{Performance on Forest Cover Dataset}
\label{tab:local_lrn}
\end{table}
As is evident from a review of Table \ref{tab:local_lrn}, a local learning approach provides better results for the forest cover dataset.

\subsection{Time Series Prediction with Big Dataset}

\cite{das2015fast} illustrate an application of the ensemble technique to a dataset from the utility domain  that has nearly two million data instances, see \cite{power_consumption_dataset} for further detail of the dataset. The dataset represents electricity consumption data at one-minute sampling intervals from a household. The dataset provides the voltage readings and three sub-meter readings for the household over a four year period. The dataset exhibits periodicity and seasonality. To capture these effects temporal features are created. The learning task is to predict the voltage from the other features. \cite{das2015fast} use bootstrapped samples to develop a Gaussian Process model for the regression task. This is very similar to the use of a random sample to develop a tree model in \cite{breiman2001random}. However, a Gaussian Process model is developed using the sample in \cite{das2015fast}. Kernel selection for the Gaussian Process regression model is determined by  using data exploration and experimental evaluation. As discussed in \cite{das2015fast}, in many problems it is observed that data usually lie in a manifold \cite{yang2015minimax} or depend on a small number of features. In such problems it is possible to develop reasonable models with a relatively small sample (see \cite{das2015fast}[Section 3] for details). The sample size is a critical parameter of the algorithm used in \cite{das2015fast}. Guidelines for picking the sample size are provided in \cite{das2015fast}[Section 3]. Ensemble methods can be combined to enhance prediction performance by using model stacking, see e.g.,  \cite{wolpert1992stacked}. Given a set of ensemble methods, model stacking picks the best performing model for a particular input. For example, the performance of the xgboost and bagged Gaussian Process models are provided in Table~\ref{tab:ens_learn}.

\begin{table}[H] 
	\centering
	\begin{tabular}{|c|c|c|}
		\hline
		BGP & XGBoost & Stacked \\ 
		\hline
		 2.09 & 1.63 & 1.61 \\
		\hline	
	\end{tabular}
\caption{RMSE (Volts) for Ensemble Learning on Household Power Consumption Data}
\label{tab:ens_learn}
\end{table}




\section{Summary}\label{sec:summary}
This paper has  provides a comprehensive  review of SML. Statistical learning theory provides the concepts and the framework to analyze machine learning algorithms. Many important  building blocks of Statistical learning theory have been motivated and presented. An overview of big data and the challenges they present to machine learning has also been provided.

Theoretical ideas from statistics and computer science are both needed to design, implement and evaluate machine learning applications. Application of machine learning is inherently computational. Computational approaches for applying machine learning to big datasets are motivated using  ideas from statistical learning theory.
We conclude by saying that success in applying machine learning  for big data systems requires:
\begin{itemize}
\item Leveraging prior experience gained either first hand personally or through careful review of literature.
\item Data of sufficient quantity and quality.
\item An understanding of the statistical and computational tools required to design algorithms, implement them and interpret the results. 
\end{itemize}




\section*{Acknowledgements}
Sourish Das's reserach has been  supported by an Infosys Foundation Grant and a TATA Trust grant to CMI and also 
 by a UK Government funded Commonwealth-Rutherford Scholarship.  

\bibliographystyle{imsart-nameyear}
\bibliography{mlbd}

\end{document}